\documentclass{article}
\usepackage{ijcai24}

\usepackage{times}
\usepackage{soul}
\usepackage{url}
\usepackage{color}      
\usepackage[dvipsnames, svgnames, x11names]{xcolor}
\xdefinecolor{tsinghua}{rgb}{0.455,0.204,0.506}
\usepackage[colorlinks,
            linkcolor=blue,
            anchorcolor=blue,
            citecolor=blue,
            backref=page
            ]{hyperref}
\usepackage[utf8]{inputenc}
\usepackage[small]{caption}
\usepackage{graphicx}
\usepackage{amsmath}
\usepackage{amssymb}
\usepackage{amsthm}
\usepackage{booktabs}
\usepackage{algorithm}
\usepackage{algorithmic}
\usepackage[switch]{lineno}
\usepackage{marvosym}

\newtheorem{definition}{Definition}

\title{A Comprehensive Survey of Cross-Domain Policy Transfer for Embodied Agents}

\author{
Haoyi Niu$^1$
\and
Jianming Hu$^{1*}$
\and
Guyue Zhou$^{1*}$
\And
Xianyuan Zhan$^{1,2}$\thanks{Corresponding authors. We will keep updating a collection of research papers on this topic in \href{https://github.com/t6-thu/awesome-cross-domain-policy-transfer-for-embodied-agents}{https://github.com/t6-thu/awesome-cross-domain-policy-transfer-for-embodied-agents}.}
\affiliations
$^1$Tsinghua University,
$^2$Shanghai Artificial Intelligence Laboratory \\
\emails
nhy22@mails.tsinghua.edu.cn, 
zhanxianyuan@air.tsinghua.edu.cn
}

\begin{document}

\maketitle
\begin{abstract}
The burgeoning fields of robot learning and embodied AI have triggered an increasing demand for large quantities of data.
However, collecting sufficient unbiased data from the target domain remains a challenge due to costly data collection processes and stringent safety requirements. 
Consequently, researchers often resort to data from easily accessible source domains, such as simulation and laboratory environments, for cost-effective data acquisition and rapid model iteration.
Nevertheless, the environments and embodiments of these source domains can be quite different from their target domain counterparts,
underscoring the need for effective cross-domain policy transfer approaches.
In this paper, we conduct a systematic review of existing cross-domain policy transfer methods.
Through a nuanced categorization of domain gaps, we encapsulate the overarching insights and design considerations of each problem setting. We also provide a high-level discussion about the key methodologies used in cross-domain policy transfer problems.
Lastly, we summarize the open challenges that lie beyond the capabilities of current paradigms and discuss potential future directions in this field.

\end{abstract}

\begin{figure*}[t]
    \centering
    \includegraphics[width=\textwidth]{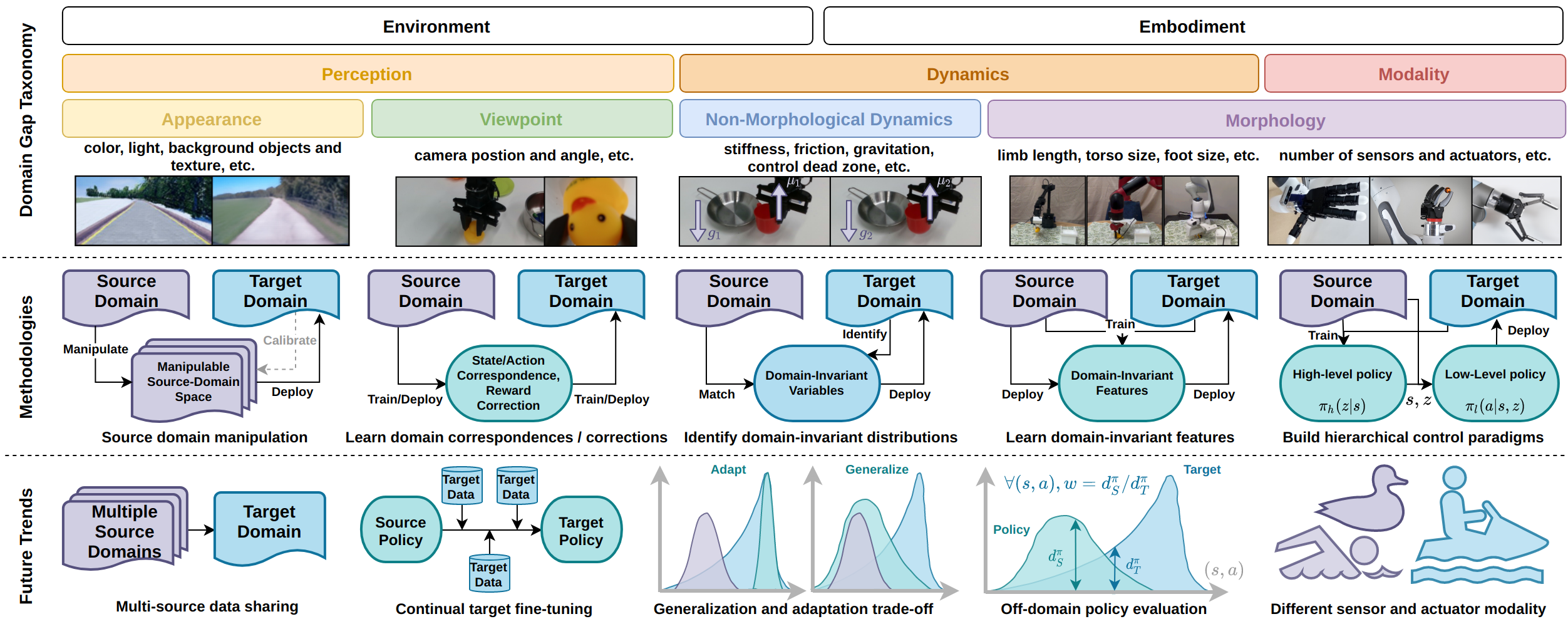}
    \caption{The main architecture of the survey: domain gap taxonomy, overarching insights on methodologies, and future trends.}
    \label{fig:cat_gap}
\end{figure*}

\section{Introduction}
The past few years have seen rapid progress
in the fields of robot learning and embodied AI, integrating advances in computer vision, decision-making, and even language processing to build capable embodied agents~\cite{duan2022survey,vuong2023open}.
This naturally leads to a surge in demand for high-quality and large-scale training data. However, collecting large amounts of data in the \textbf{target domain} (\textit{where the policy is deployed, i.e., the real world/task environment}) at will can be prohibitively costly due to efficiency issues and safety concerns, 
e.g., in autonomous driving and industrial robot control~\cite{chen2023end,nguyen2011model}.
Instead, a popular practice is to utilize data from easily accessible \textbf{source domains} 
(\textit{e.g., simulation or laboratory environments})
that allow safe exploration and cheap data collection.
Despite the advances in simulation modeling technologies, high-fidelity simulators still struggle to capture nuanced physical properties, as well as delicate environmental and embodiment details of target domains~\cite{tobin2017domain,peng2018sim}.
In some settings, although human demonstration videos can be easily recorded in a controllable manner in the target environment, the distinct embodiment from the target robot agents hinders their direct use in policy learning~\cite{yu2018oneshot}. 
Such intricate environment and embodiment discrepancies, also referred to as \textit{domain gaps}, negatively impact policies trained on source domain data and inevitably lead to their deployment failures in the target domains.
The data bottlenecks in real-world tasks and the wide existence of domain gaps naturally stimulated cross-domain policy transfer studies, aiming to fully exploit existing off-domain data to learn transferable policies.

Cross-domain policy transfer has emerged as a crucial class of methods for erasing the influences of domain gaps on policies, serving as a bedrock for large-scale real-world deployment of embodied agents~\cite{chebotar2019closing,bewley2019learning}.
However, existing approaches in this direction are highly fragmented, primarily due to diverse types of domain gaps, various learning paradigms, as well as distinct data constraints and setting assumptions.
Such fragmentation seriously shadows our understanding of the underlying connections and differentiations among existing policy transfer strategies, making it difficult for researchers to gain a holistic view of this field and embark on new research endeavors.

In light of this, 
we present the first comprehensive review of cross-domain policy transfer methods for embodied agents.
We begin by unifying the related notations and definitions in cross-domain settings, based on which we provide clear categorization of different types of domain gaps, with discussion on their distinctions and connections.
Then, we consolidate a vast diversity of existing methods,
organizing them into four commonly encountered domain gap categories in the literature, i.e., appearance, viewpoint, dynamics, and morphology gaps.
Furthermore, we provide overarching methodological insights developed in existing approaches, 
and discuss open challenges and promising future research directions. 
A detailed architecture of the survey, including domain gap taxonomy, methodology classification, and future trends are illustrated in Figure~\ref{fig:cat_gap}. We hope our survey can bring new insights and expedite future research in cross-domain policy transfer for embodied agents.

\section{Overviews}
\subsection{Notations and Definitions}
\begin{definition}[Domain]
    With environment $\mathcal{E}_\text{env}$, embodiment $\mathcal{E}_\text{emb}$, and the embodiment dynamics influenced by both properties $T(\mathcal{E}_{\text{env}}, \mathcal{E}_{\text{emb}})$, we can define a domain $\Omega=\{\mathcal{E}_{\text{env}}, \mathcal{E}_{\text{emb}}, T(\mathcal{E}_{\text{env}}, \mathcal{E}_{\text{emb}})\}$.
\end{definition}
\begin{definition}[Domain-Dependent Markov Decision Process (D-MDP)]
Adapting standard MDP with domain-dependent information
as $\mathcal{M}(\Omega):=\langle\mathcal{S}_\Omega, \mathcal{A}_\Omega, T_\Omega, R_\Omega, \gamma\rangle$ where $\mathcal{S}_\Omega$, $\mathcal{A}_\Omega, T_\Omega$ are domain-dependent state, action spaces, and dynamics; $R_\Omega$ is the task-relevant reward function also influenced by $\mathcal{S}_\Omega$ and $\mathcal{A}_\Omega$; $\gamma$ denotes the discount factor.
\end{definition}
To keep the notations uncluttered,
we denote $\Omega_{src}=\{\Omega_{src}^{i}\}^N_{i=1}, N\in\mathbb{N}^+$ as the source domain(s), $\Omega_{tgt}$ as the target domain, and use subscript $src$ and $tgt$ to indicate elements from source and target domains in the rest of the paper. 
Next, we introduce the definition for cross-domain policy transfer:
\begin{definition}[Cross-Domain Policy Transfer]
Given a policy class $\Pi$, one or multiple policies $\pi_{src}\in\Pi$ obtained in $\Omega_{src}$, and an explicit or implicit policy transfer method $h: \Pi\rightarrow\Pi$, an effective cross-domain policy transfer is achieved when $J_{tgt}(\pi_{src}) \ll J_{tgt}(h(\pi_{src})) \leq J_{tgt}(\pi_{tgt}^*)$, where  $\pi_{tgt}^*\in\Pi$ is the optimal target policy and $J_{tgt}(\cdot)$ is some policy performance evaluation metric based on the criteria (e.g., expected cumulative rewards, success rate, etc.) of target domain $\Omega_{tgt}$.
\end{definition}



\subsection{Categorization of Domain Gaps}
As indicated in the above definitions, domain gaps arise from both the inconsistencies in environments~\cite{tobin2017domain,peng2018sim,chebotar2019closing} and embodiments~\cite{gupta2021embodied}.
In the following, we establish fine-grained classifications of domain gaps and discuss their connections as well as distinctions. 

In terms of environment inconsistencies, \textbf{appearance gaps} arise when observations in the source domain (e.g., simulations) exhibit differences in colors, background objects, illumination conditions, and rendering textures as compared to the target domain (e.g., reality), such as variations in coarse and fine rendering or high and low resolutions~\cite{tobin2017domain,andrychowicz2020learning}. Additionally, the configuration of sensor setups (e.g., camera position and angles, etc.) can significantly influence the downstream policy learning of embodied agents, 
we refer to these as \textbf{viewpoint gaps}~\cite{Sermanet2017TCN}. 
Appearance and viewpoint gaps are sometimes jointly termed \textit{visual gaps} that systematically distort the state space $\mathcal{S}_{src}$ in the source domain, or more generally, \textbf{perception gaps}~\cite{wang2022versatile} in cases where observations come from other perception sensors besides visual cameras, such as lidars.

At the intersection of embodiment and environment variations, \textbf{dynamics gaps}~\cite{peng2018sim,eysenbach2020off,niu2022when} occur when interactions between embodiments and their deploying environments, or interactions among different parts of the embodiment itself, follow different transitional dynamics ($T_{src}\neq T_{tgt}$), such as stiffness, gear dead zones of embodiments, body mass, and friction. 
Focusing on the embodiment aspect, \textbf{morphology gaps}~\cite{hejna2020hierarchically,gupta2021metamorph} arise when target embodiments exhibit different morphological designs compared to the source domain agents, e.g., variations in joint types, module shapes, and lengths, which may ultimately lead to a dynamics mismatch. Morphological differences may also encompass variations in the dimensions and semantic meanings of state and action spaces $\mathcal{S}_{src}, \mathcal{A}_{src}$, such as the number of observational sensors, limbs, and end effectors. In some literature, these are referred to as \textbf{modality gaps}~\cite{wang2022weakly,salhotra2023learning} for different sensing and actuation modalities.
Occasionally, researchers merge the gaps resulting from morphology discrepancies and parts of non-morphological dynamics disparities, which stem from internal physical properties, into a unified term: \textit{embodiment gaps}. This term represents a more ego-centric robotic perspective, independent of external environmental variations, and should be clearly distinguished from morphology gaps.

\section{Policy Transfer Across Different Gaps}
\subsection{Cross-Appearance Policy Transfer}
To remedy the appearance gap, a class of unsupervised transfer learning techniques originating from visual domain adaptation has been proposed. 
These techniques map observational representations from the source domain (e.g., simulations) to the target domain (real world) while ensuring consistent data distributions. 
This growing collection of approaches is particularly adept at image-to-image translation across domains, making the observation information transferable for end-to-end vision-based autonomous robots and vehicles~\cite{triess2021survey}. 

Various schemes for the mapping function have been proposed, addressing the problem from different perspectives.
Some studies adopt the cycle consistency philosophy from CycleGAN~\cite{zhu2017unpaired} to ensure a photo-realistic image translation process, achieving good real-world transfer performance with only a modest number of real-world observations~\cite{rao2020rl}. 
Conversely, real-to-sim translation~\cite{zhang2019} requires a pre-trained adaptation module to convert real images captured by cameras into simulation-like images at test time, which can be computationally inefficient during real-world deployment.
As an alternative, training a canonical domain-invariant representation, such as semantic segmentation~\cite{Pan2017b,mueller2018driving,wang2022versatile}, enables observations from both domains to be translated into an intermediate and lower-dimensional representation. 
This unifies the semantic meaning of the observation space while easing the burden of the downstream policy learning module. 
In contrast to explicit representations like semantic segmentation, intra-domain image reconstruction with direct and cyclic losses~\cite{bewley2019learning} offers another way to enhance transferability, where a bi-directional image translation strategy is introduced to form an implicit structure of representation.

Domain randomization and visual data augmentation~\cite{tobin2017domain,laskin2020reinforcement,kar2019meta,yue2019domain} instead opt for a domain generalization approach that focuses on manipulating pixel-level physical mechanisms. These methods do not require massive data from the target domain or learning transferable embeddings.
Additionally, proper model setups can also enhance the policy feasibility for real-world execution. For instance, interactive imitation learning~\cite{lee2022beyond} is specially tailored in simulation to distill state-based experts into a ``student" vision-based policy, allowing for in-domain data augmentation from randomized simulations.

\subsection{Cross-Viewpoint Policy Transfer}
In many cases, training data may not always be available from a first-person or ego-centric viewpoint, which is often the most convenient and desirable observational input~\cite{pathak2018zeroshot}. 
Embodied agents often have different camera setup positions and angles, resulting in observational information with systematic bias. 
Policy learning that relies on robust cross-appearance visual encoders can still be vulnerable to viewpoint discrepancies, such as a wrist camera on source-domain demonstrators and a side camera in the target environment. 
To address this, agents need to translate (imagine) third-person observations from their own viewpoint.

To relax the assumption in ``learning from demonstrations'' that demonstrations come solely from an identical observational configuration, third-person imitation learning~\cite{stadie2016third} constructs an architecture based on generative adversarial imitation learning, which minimizes class loss (expert vs. non-expert) while maximizing domain confusion. 
Minimizing class loss enables the model to accurately predict the correct class label for a given input, which is crucial for task completion;
maximizing domain confusion allows the model to generalize better across different domains, adapting to real-world situations where internet-scale data comes from third-person demonstrators with different viewpoints from embodied agents designated for later deployment.

In a different vein, \cite{liu2018imitation} learns a context translation model that can convert a demonstration from one context (e.g., a third-person viewpoint and a human demonstrator) to another context (e.g., a first-person viewpoint and a robot).
This approach directly predicts demonstrator behavior sequences from the target robot's viewpoint, which is claimed to excel in more complex manipulation skill acquisition. 
Another line of work~\cite{sadeghi2018sim} suggests that training a deep convolutional recurrent neural network implicitly learns to identify the effects of actions in image space from the past history of observations and actions. 
This enables robots to understand how actions affect their motion from the current viewpoint, given a small number of labeled target image queries. 
Contrastive learning~\cite{Sermanet2017TCN} is also employed to discover attributes that remain consistent across viewpoints or even change throughout task progress.

However, in situations where source domain demonstrators have not only different viewpoints but also distinct morphological embodiments, learning domain-invariant features alone may not suffice for transferrable agent learning.
To address these challenges, meta-learning approaches~\cite{yu2018oneshot} have been introduced, although models for each unseen task must be trained separately, necessitating more data and high-capacity models for generalization. 
Alternatively, a hierarchical setup has also been proposed~\cite{sharma2019third}, in which an embodiment-agnostic high-level module learns to generate first-person sub-goals conditioned on third-person demonstrations and an embodiment-specific low-level controller predicts actions to achieve those sub-goals.

\subsection{Cross-Dynamics Policy Transfer}
Embodied tasks, regardless of whether they have visual observation or not, involve complex transition dynamics that dictate interactions with the environment and constraints within the embodiment's components. 
These intricacies pose considerable challenges in building
high-fidelity simulators or finding unbiased source domains. 
Traditional system identification methods~\cite{ljung1998system,kolev2015physically,yu2017preparing} address domain inconsistency through dynamics model fitting and calibration, which often involve extensive target domain data collection and perform poorly in complex physical dynamics.

An alternative approach is to modify the source domain configuration directly, assuming access to manipulable source domains. 
Many works have attempted to randomize the physical parameter space of the source domain simulators with different configurations to improve generalization across various target domains~\cite{rajeswaran2017epopt,peng2018sim,andrychowicz2020learning}. 
However, this approach often requires manually specified, sufficiently large parameter spaces for adjustment~\cite{vuong2019pick}, which can be impractical for complex embodied systems. 
Active domain randomization~\cite{mehta2020active} addresses the sample complexity issue by adaptively selecting parameters from the most informative configurations according to the discrepancies of policy rollouts in randomized and reference environment instances.
Another perspective is to revisit classical system identification and lower its demand for target data, 
such as incorporating target-domain prior information to guide accurate and efficient source-domain posterior distribution calibration~\cite{muratore2021data,ramos2019bayessim,tan2016simulation,du2021auto}. 
Grounded action transformation (GAT)~\cite{hanna2017grounded} learns target-domain forward dynamics models and adjusts source-domain inverse dynamics accordingly, modifying source dynamics to better match target dynamics.

When source domains are not white-box or modifiable, many recent dynamics adaptation approaches focus on regularizing \textit{policy learning} rather than \textit{dynamics modeling}, assuming a fixed source domain. GARAT~\cite{desai2020imitation} learns an adversarial imitation-from-observation policy by discriminating between generated actions and target environment actions, bypassing the need for a modifiable source domain. 
DARC and related methods~\cite{eysenbach2020off,liu2021dara} solve cross-dynamics reinforcement
learning (RL) via reward correction to compensate for dynamics shifts across domains in online or offline settings. 
H2O and H2O+~\cite{niu2022when,niu2023h2o+} introduce a dynamics-aware hybrid offline-and-online RL paradigm, integrating learning from online simulation and offline real-world data in a single-stage learning process while correcting dynamics gaps during policy learning. 
VGDF~\cite{xu2023crossdomain} samples source domain transitions $(s_{src}, a_{src}, s'_{src})$ with small value difference between $s'_{src}$ and $s'_{tgt}$ (obtained from a learned target domain dynamics model), and combines the selected source domain data and target domain counterparts for policy learning.

To address dynamics gaps more affordably, some approaches harness task-relevant, domain-agnostic information in state transitions.
SAIL~\cite{Liu2020State} advocates for state alignment in cross-domain imitation learning (IL), as optimal policies heuristically induce similar state trajectories under different imitator and expert dynamics. 
SAIL enforces global state distribution matching based on Wasserstein distance and local state transition alignment based on $\beta$-VAE. 
Concurrent work~\cite{Gangwani2020State-only} leverages the Wasserstein distance of state visitation distributions from both domains and an adversarial IL paradigm for policy optimization. 
Additionally, incorporating an inverse dynamics policy learned with target demonstrations~\cite{christiano2016transfer} offers an alternative for matching (next-)state distributions. 
HIDIL~\cite{jiang2020offline} extends this idea with Horizon-Adaptive Inverse Dynamics, matching states from both domains in an H-step horizon and recovering feasible actions in the target domain based on the inverse dynamics policy. 
SOIL~\cite{radosavovic2021state} and SRPO~\cite{xue2023state} further develop these ideas, with the latter extending this insight to RL and providing theoretical grounding for the assumption that optimal policies under different dynamics induce similar stationary state visitation distributions.

However, the assumption of identical state reachability in source and target domains does not always hold in real-world situations. Feasibility MDP (f-MDP)~\cite{cao2021learningfea,cao2021learningfrom} addresses this issue by calculating feasibility scores to weigh the learning signal of each demonstration. Cold diffusion techniques can also be adapted for feasibility-guided trajectory planning by degrading every state in source trajectories to the nearest recorded state in the target replay buffer~\cite{wang2023cold}.
\subsection{Cross-Morphology Policy Transfer}
Morphology gaps typically only affect low-level control, which naturally favors a decoupled hierarchical solution with a morphology-specific low-level policy and a transferable high-level policy~\cite{hejna2020hierarchically}.
The high-level policy takes morphology-agnostic state observations and generates sub-goals for the low-level policy to follow. 
MAIL~\cite{salhotra2023learning} leverages domain-invariant features in the observation space, such as end effector positions, as optimal position trajectories should be task-relevant and morphology-independent. 
It performs 
position-based matching at the high level and uses inverse dynamics to recover morphology-specific low-level action commands. Alternatively, TAME~\cite{hejna2020task} explores joint optimization for the best morphology design that benefits the embodiment for successful task execution, and optimal policies corresponding to this morphology. 

Morphology can also be considered as another modality that can be conditioned on models with the Transformer architecture~\cite{vaswani2017attention}.
Morphology-aware Transformer~\cite{yu2023multi} captures meaningful patterns between robot embodiment and actions using a causally masked Transformer, allowing conditional action generation based on desired robot embodiment, past states, and past actions. 
However, learning a universal controller for a population of morphologies is resource-intensive and infeasible due to the exponentially increasing morphology representation space~\cite{gupta2021embodied}.
Focusing on learning embodiment-dependent policies, MetaMorph~\cite{gupta2021metamorph} first encodes morphology representation into a vector sequence, concatenates it with proprioceptive position embeddings, and processes it using a morphology-aware Transformer. 
The Transformer output is successively concatenated with exteroceptive observations before passing through the final action decoder. 
Large-scale pre-training over libraries of different morphologies is also utilized to facilitate sample-efficient transfer to new robot morphologies and tasks.

\subsection{Cross-Multi-Gap Policy Transfer}
In many complex tasks, we might simultaneously encounter multiple types of domain gaps due to substantially different embodiments and deployed environments. 
However, most previous works focus on paired and temporal-aligned data from source and target domains, and only address a certain type of domain gaps, which limits their applicability in general settings unless extra designs are introduced.

From a general perspective, correspondence learning across domains can empirically align the properties of both source and target domains by constructing direct mappings.
GAMA~\cite{kim2020domain} learns state and action correspondence mapping $f:\mathcal{S}_{src}\rightarrow\mathcal{S}_{tgt}, g: \mathcal{A}_{tgt}\rightarrow\mathcal{A}_{src}$ from unpaired and unaligned demonstrations, and then adapts source policy $\pi_{src}$ to feasible target policy as $\pi_{tgt}=g\circ\pi_{src}\circ f$.
To jointly optimize state and action correspondence models, 
adopting dynamics cycle consistency~\cite{zhang2020learning} allows for splitting 
action mapping $g$ into coupled dual mappings $p: \mathcal{S}_{src}\times\mathcal{A}_{src}\rightarrow\mathcal{A}_{tgt}$ and $q: \mathcal{S}_{tgt}\times\mathcal{A}_{tgt}\rightarrow\mathcal{A}_{src}$, which builds correlations between action and state correspondences.
For various scenarios that only involve suboptimal policies as demonstrators, such as internet videos of humans performing tasks,
\cite{raychaudhuri2021cross} enforces cycle consistency on the state space together with a normalized position estimator function to align trajectories across domains without the need for expert actions.
To handle the remaining domain misalignment issue of adopting unsupervised cycle consistency techniques, WeaSCL~\cite{wang2022weakly} introduces weak supervision into correspondence learning with temporal ordering and paired abstraction data.

Instead of learning direct correspondences, an emerging avenue of studies has extended learning domain-invariant features from closing appearance and viewpoint gaps to simultaneously addressing other domain gaps, e.g., dynamics and morphology gaps.
From internet-scale cross-embodiment videos, XIRL~\cite{zakka2021xirl} leverages temporal cycle consistency to ensure task-progress aware and domain-agnostic representation learning so that distance from goal state represented in that embedding space can be regarded as rewards used for policy training on novel embodiments.
Motivated by abstracting task information from state space to ease the burden of downstream policy transfer, \cite{franzmeyer2022learn} proposes a mutual information criterion to reduce target state space with mapping $f': \mathcal{S}_{tgt}\rightarrow\mathcal{Z}$ to a task-relevant domain-invariant embedding $\mathcal{Z}$, and then jointly learning source mapping $g': \mathcal{S}_{src}\rightarrow\mathcal{Z}$ and an adversarial imitation policy that generates state transitions closely resembling the expert target state transitions.
From a more robotic perspective, skill acquisition is an explicit procedure for grounding task-specific domain-agnostic features for easy transfer.
With paired data from both domains, agents learn multiple skills and transfer knowledge by training in invariant feature spaces, upon which target domain agents can acquire new skills mastered by source domain agents~\cite{gupta2017learning}.
In a hierarchical scheme, STAR~\cite{pertsch2022cross} pre-trains a low-level policy to decode actions from learned high-level semantic skill policies that select a transferable skill in target task learning.
With unpaired and unaligned cross-embodiment videos, XSkill~\cite{xu2023xskill} pre-trains skill discovery models for further skill identification. In XSkill, a skill alignment transformer is introduced to detect, align, and compose the learned skills to complete new tasks, and then pass the inferred skills to a skill-conditioned diffusion policy to output the robot's actions.

Additionally, contrastive learning also offers a general and natural solution for aligning domain representation with positive and negative samples, from large amounts of in-the-wild cross-embodiment unpaired data.
Polybot~\cite{yang2023polybot} aligns observation and action spaces using an engineering approach and then aligns policy's internal representations through contrastive learning to combat other domain discrepancies.
Based on prompt-based learning, CONPE~\cite{choi2023efficient} develops a novel contrastive prompt ensemble framework that uses the CLIP vision-language model~\cite{radford2021learning} as the visual encoder and facilitates dynamic adjustments of visual representations against domain changes through an ensemble of contrastively learned visual prompts.
VIP~\cite{ma2022vip} contrastively pre-trains an (implicit) visual goal-conditioned value function that aims to capture task-agnostic goal-oriented representations, which can generalize to unseen domains and tasks.
As a multi-modal extension of VIP, LIV~\cite{ma2023liv} learns vision-language representations from language-annotated videos.
In a similar setting, DecisionNCE~\cite{li2024decisionnce} learns universal embodied multimodal representations through an infoNCE-style learning objective, derived based on reward reparameterization under the preference-based learning framework.
RT-X~\cite{vuong2023open} harnesses domain alignment and scene understanding ability of large vision-language models, unifying domain representations with domain-invariant task-relevant language instructions.

\section{Discussions on Methodologies}
\subsection{Source Domain Manipulation}
Modifying the source domain modeling is undoubtedly the most straightforward solution to close the domain gaps when source domains (e.g., simulators) are manipulable. Such modifications often include: randomizing partially known or modifiable modeling configurations for better generalization in target domains~\cite{tobin2017domain,rajeswaran2017epopt,peng2018sim,lee2022beyond,mehta2020active,andrychowicz2020learning}, model calibration to match better with the target domains~\cite{kolev2015physically,tan2016simulation,yu2017preparing,hanna2017grounded,ramos2019bayessim,muratore2021data,du2021auto}, and adapt morphological design configurations from libraries of candidates~\cite{hejna2020task,gupta2021embodied,gupta2021metamorph,yu2023multi}.
Large-scale source-domain randomization has been widely applied to visual properties (i.e., color, texture, lighting condition, shapes and types of interactive objects, camera position, orientation, and field of view)~\cite{tobin2017domain}, dynamics parameters (i.e., mass, damping, friction, and control timestep)~\cite{peng2018sim}, or oftentimes both~\cite{lee2022beyond}. 
With principled morphology representation, we could effectively manipulate action space, sensory inputs, module shape, and size so that agents can generalize in the vast modular morphology design space.
This promotes domain generalization ability maximally with large-scale parallel computing resources to support high-dimensional parameter randomization in simulation, which is highly recommended in industrial practice~\cite{lee2022beyond}.

In cases when access to target domain data is allowed but suffers costly and laborious collection, source-domain calibration turns out to be fundamental, direct, and effective for simple and well-modeled environments~\cite{yu2017preparing}.
However, target domain environments can be quite complex in practice, even strong calibration approaches are likely to under-model intricate target domain physical properties and procedures, such as non-rigidity, gear dead zone, wear-and-tear, and rolling friction.
Such properties are hard to capture in current simulation modeling technologies.
To summarize, source domain manipulation is highly dependent on manipulable source domains, nuanced and comprehensive knowledge of environment modeling, and rich computation resources, as preferred by the industry field.

\subsection{Learn Domain Correspondences / Corrections} 
Learning mapping functions (correspondences) or correction terms is another class of commonly used techniques to handle domain gaps.
Image-to-image translation~\cite{zhu2017unpaired,Pan2017b,zhang2019,rao2020rl} build mappings between visual representations across domains. 
The viewpoint context translation model~\cite{liu2018imitation} converts data from the source context to ones from the target viewpoint.
Unlike previous works, learning state and action correspondences for domain alignment~\cite {kim2020domain,zhang2020learning} allows for direct leverage of unpaired data.
Additional designs can also be added in this general framework to
reduce the need for expert action collection~\cite{raychaudhuri2021cross}, which fits well with the common setting of learning from action-free video demonstrations.
To address the accuracy issue of correspondence learning under stricter conditions, WeaSCL~\cite{wang2022weakly} finds a trade-off between strong supervision of strictly paired data and regularization over unpaired data.
In addition to learning correspondences, learning reward correction for domain gap compensation~\cite{eysenbach2020off,liu2021dara,xue2023state} and dynamics ratio for reweighting learning signals on source domain samples~\cite{niu2022when,niu2023h2o+} have also become viable practices in cross-domain RL.
\subsection{Identify Domain-Invariant Distributions}
Identifying domain-invariant distributions from accessible and organized data offers a seemingly simpler solution for cross-domain policy transfer without extracting detailed domain-dependent correspondences.
For example, state regularization in IL~\cite{christiano2016transfer,Liu2020State,Gangwani2020State-only,jiang2020offline,radosavovic2021state} and RL~\cite{xue2023state} develop upon the insight that optimal policies across different domains induce similar state visitation distribution. 
Cold diffusion is used in diffusion-based planning to constrain the generated states inside the state distribution of the target replay buffer~\cite{wang2023cold}.
In some cases, partial state distribution matching (e.g., position-based matching) is adopted since not all dimensions of the state space are semantically identical across domains~\cite{salhotra2023learning}.
However, this avenue of work typically only addresses dynamics and morphology gaps without discrepancies in visual representation; otherwise, extra cross-domain correspondences/representations are required to align the state space before performing state-based matching.
Additionally, extra efforts are needed to apply these works to long-horizon compositional tasks since state distribution matching lacks task-progress awareness.
\subsection{Learn Domain-Invariant Features}
Learning task-relevant domain-invariant representations is also a principled and popular direction to bridge domain gaps~\cite{stadie2016third,Sermanet2017TCN,mueller2018driving,bewley2019learning,zakka2021xirl,franzmeyer2022learn,wang2022versatile}, which sometimes also appear in the form of skills~\cite{gupta2017learning,pertsch2022cross,xu2023xskill} and sub-goals~\cite{sharma2019third}.
As canonical representation across domains can be reused in multiple and even new target contexts, offering great flexibility and data efficiency, however, these approaches could also suffer some barriers as compared to 
learning correspondences.
The domain-invariant representations, for instance, could require additional efforts for tackling issues like uninformative degenerated mapping~\cite{gupta2017learning}.
Furthermore, this line of works often solely focuses on learning invariant features in observations,
unlike learning state and action correspondences that could seamlessly align different MDPs temporally, which brings better task progress awareness for planning tasks.
This highlights the extra need to borrow off-the-shelf temporal vision alignment techniques for pairing demonstrations, e.g. temporal contrastive network~\cite{Sermanet2017TCN} and temporal cycle consistency~\cite{dwibedi2019temporal}.
However, with self-supervision on comprehensive long-horizon multi-skill demonstrations, the learned representations sometimes could also be progression/distance-aware, holding advantages of yielding goal-directed visual reward~\cite{Sermanet2017TCN,zakka2021xirl,ma2022vip,li2024decisionnce} and guiding the downstream policy optimization process.
It also provides possibilities for incorporating language instructions into unified vision-language cross-embodiment representations~\cite{ma2023liv,li2024decisionnce}.






\subsection{Build Hierarchical Control Paradigms}\label{hierarchical}
The essence of hierarchical frameworks in cross-domain settings is to decouple the action output procedure into domain-independent high-level policy learning (e.g., skill aquisition~\cite{gupta2017learning,pertsch2022cross,xu2023xskill} and sub-goal generation~\cite{hejna2020hierarchically,sharma2019third}) and domain-specific low-level policy learning. Such a treatment greatly reduces the difficulties in domain gap modeling and has proven to excel in solving many complex tasks.
In a similar philosophy, some meta-learning methods~\cite{finn2017model,yu2018oneshot,nagabandi2018learning,zintgraf2019varibad,rakelly2019efficient} train context-based hidden embeddings for fast adaptation, with which the meta-learned policy can adapt to target environments by fine-tuning on a small amount of target domain data.
\section{Open Challenges and Future Trends}\label{challenges}
\subsection{Different Sensor and Actuator Modalities}
The transferable knowledge from source domain embodiments can be quite limited when dealing with significantly different state and action modalities~\cite{wang2022weakly,salhotra2023learning}. Recently, large and expressive foundation models, combined with large-scale data collected from diverse robotic tasks, have emerged as a promising direction for cross-embodiment policy transfer~\cite{vuong2023open}. Octo~\cite{octo_2023}, a transformer-based diffusion policy model, serves as a versatile policy initialization that can be effectively fine-tuned to adapt to new observation and action spaces. Its block-wise attention structure allows for adding or removing new inputs and outputs with various modalities as needed. However, it remains unclear what and how an agent can learn from data from significantly different embodiments of the same task.

Most studies in cross-modality settings heavily rely on expert demonstrations, which are costly to collect and limited in size,
causing the issue that expert policies are hard to learn or transfer across modalities.
Future work could focus on learning effective and transferable information from non-optimal, in-the-wild demonstrations to address these limitations.

\subsection{Multi-Source Data Sharing}
Current research on learning correspondences between domains typically focuses on narrow settings, where data are assumed to originate from only two domains~\cite{stadie2016third,bewley2019learning,niu2022when}. 
However, in practice, it is crucial to handle multiple source domains to overcome the data scarcity issue~\cite {xue2023state}. 
Modern cross-domain methods need more flexible interfaces to incorporate multi-source data with domain gaps of varying scales.

A versatile and expressive feature space could also be developed to unify the representation of data across different domains. 
In addition to focusing on representation learning, another potential direction is to directly filter or edit source data according to learned criteria in the target domain. 
This perspective emphasizes manipulating data as a means to address the challenges of multi-source data sharing.
\subsection{Continual Target Fine-Tuning}
Current cross-domain policy transfer approaches often lack flexible designs to accommodate various forms of source-domain information, such as data and pre-trained policy. 
Sometimes, we might desire to fine-tune the source-domain policy using target data, as
target data might not be readily available at the beginning of training and only obtainable sporadically, encompassing different coverages and skillsets. 
This necessitates a policy model that is compatible with continual learning. 
Essentially, it could also help relax the long-standing assumption that the target domain remains time-invariant, while real-world systems often deviate from this due to factors such as wear and tear.

A potential approach to leverage multi-stage target data is the adoption of continual fine-tuning ~\cite{li2023proto,smith2023grow}. 
These techniques enable the continuous integration of target data for fine-tuning, thereby facilitating continual skill acquisition and policy adaptation in time-varying target domains. 
As a result, the development of a generalist, multi-skill, multi-domain policy after fine-tuning becomes more achievable, paving the way for more robust and versatile cross-domain solutions in future research.

\subsection{Generalization and Adaptation Trade-Off}
Current cross-domain transferable policies tend to be either highly adaptive to a single accessible target domain or moderately generalizable to various random domains. Striking a balance between generalization and adaptation, akin to the trade-off between generalists and specialists in a cross-domain setting, appears to be a challenging task. To find the nuanced equilibrium, employing powerful foundation models and extensive cross-domain data could serve as a practical solution to such demanding requirements~\cite{reed2022a}.

Recently, we have witnessed the burgeoning emergence of large (vision-)language models, which can naturally serve as powerful domain aligners since language is fundamental, easily attainable, information-abstract, and domain-transferable~\cite{vuong2023open,ma2023liv,li2024decisionnce}. Moreover, language models have the potential to serve as domain generalizers due to their strong common sense reasoning abilities across a wide range of everyday scenarios. However, vision-language representation alignment remains a longstanding challenge, as we cannot expect language models to either generalize to or adapt to desired target domains without resolving the alignment of representations with diverse input modalities.
\subsection{Off-Domain Policy Evaluation}
Evaluating policies in target domains can sometimes be prohibitively expensive and even hazardous, while continuous access to source domains allows for faster and more controlled policy evaluation, albeit with reduced reliability due to domain gaps. 
However, there is a scarcity of theoretical or principled criteria to determine whether a policy model evaluated in source domains can be successfully transferred to target domains~\cite{katdare2023marginalized}. 
Therefore, future research should develop reliable \textit{off-domain policy evaluation} methods (an extension of off-policy evaluation under domain gaps) together with standardized real-world benchmarks~\cite{walke2023bridgedata,vuong2023open}. 
These efforts are expected to offer a principled procedure and rigorous criteria for evaluating the transferability of policies using accessible source domains and limited pre-collected target data, which are the common settings in real-world scenarios.




\section{Conclusion}

In this survey, we provide the first comprehensive review of the rapidly evolving field of 
cross-domain policy transfer for embodied agents.
We have unified the notations and definitions in cross-domain settings, distinguishing various types of domain gaps and clarifying their connections and differences. By categorizing the highly fragmented approaches in the literature, we shed light on the methods used to address appearance, viewpoint, dynamics, and morphology gaps.
Moreover, we provide overarching insights shared among these methodological approaches and discuss open challenges as well as promising future trends.
As the field of embodied AI continues to evolve, addressing these challenges and embracing emerging trends will be crucial for developing more robust and versatile solutions for real-world deployment. We hope our survey can serve as a useful tool for 
future research, offering a clear understanding of the current state of cross-domain policy transfer and providing a roadmap for tackling the remaining challenges in this exciting and rapidly growing field.




\section*{Acknowledgments}
This work is supported by National Natural Science Foundation of China under Grant No. 62333015 and No. 62133002, National Key Research and Development Program of China under Grant (2022YFB2502904), Beijing Natural Science Foundation L231014, and funding from Haomo.AI.

\bibliographystyle{named}
\bibliography{ijcai24}

\end{document}